\documentclass[11pt,a4paper]{article}
\usepackage{acl2015}
\usepackage{times}
\usepackage{latexsym}
\usepackage{amsmath}
\usepackage[usenames,dvipsnames]{xcolor}
\usepackage{multirow}
\usepackage{url}
\usepackage{graphicx}
\usepackage{caption}
\usepackage{subcaption}
\usepackage{ amssymb }
\usepackage{array}
\usepackage{todonotes}
\usepackage{arydshln}
\usepackage{footnote}
\usepackage{paralist}

\newcommand{\Note}[2]{} 
\newcommand{\SideNote}[2]{} 
   
\newcommand{\SideNoteMRG}[1]{\SideNote{blue!40}{#1 --MRG}}   
\newcommand{\NoteMY}[1]{\Note{green!40}{#1 --MY}}   
\newcommand{\SideNoteMY}[1]{\SideNote{green!40}{#1 --MY}}   
   
\newcommand{\SideNoteMD}[1]{\SideNote{red!40}{#1 --MD}}   

\newcommand{\secref}[1]{\S~\ref{#1}}
\newcommand{\tabref}[1]{Table~\ref{#1}}
\newcommand{\fct}{\textsc{fcm}}

\newcommand{\bx}{\mathbf{x}}
\newcolumntype{H}{>{\setbox0=\hbox\bgroup}c<{\egroup}@{}}

\newcommand{\feat}[1]{\texttt{{\scriptsize #1}}} 

\newcommand{\featnormal}[1]{\texttt{{#1}}} 
\newcommand{\vc}[1]{\boldsymbol{#1}}

\newcommand{\CommentForSpace}[1]{}

\title{Improved Relation Extraction with\\Feature-Rich Compositional Embedding Models}

\author{
Matthew R. Gormley$^{1*}$ \and Mo Yu$^2$\thanks{$^*$Gormley and Yu contributed equally.} \and  Mark Dredze$^1$\\
 $^1$Human Language Technology Center of Excellence\\
 Center for Language and Speech Processing \\
 Johns Hopkins University, Baltimore, MD, 21218 \\
$^2$Machine Intelligence and Translation Lab\\
 Harbin Institute of Technology, Harbin, China \\
\texttt{gflfof@gmail.com, \{mrg, mdredze\}@cs.jhu.edu} \\
}

\date{}

\begin{document}
\maketitle
\begin{abstract}
  Compositional embedding models build a representation (or embedding)
  for a linguistic structure based on its component word embeddings.
  We propose a Feature-rich Compositional Embedding Model (\fct{}) for relation extraction that is expressive,
  generalizes to new domains, and is easy-to-implement. The key idea is to
  combine both (unlexicalized) hand-crafted features with 
  learned word embeddings. The model is able to directly tackle the difficulties met by
  traditional compositional embeddings models, such as handling
  arbitrary types of sentence annotations and utilizing global
  information for composition.  We test the proposed model on two relation
  extraction tasks, and demonstrate that our model outperforms both 
  previous compositional models and traditional feature rich models
  on the ACE 2005 relation extraction task, and the SemEval 2010 relation
  classification task. The combination of our model and a log-linear classifier
  with hand-crafted features gives state-of-the-art results.
  We made our implementation available for general use\footnote{https://github.com/mgormley/pacaya}.
 %
\end{abstract}

\newcommand{\colorA}[1]{{\color{BrickRed}#1}}
\newcommand{\colorB}[1]{{\color{Orange}#1}}
\newcommand{\colorC}[1]{{\color{Green}#1}}

\section{Introduction}
Two common NLP feature types are lexical properties of words and unlexicalized
linguistic/structural interactions between words. Prior work on
relation extraction has
extensively studied how to design such features by combining
\emph{discrete} lexical properties (e.g. the identity of a word, its
lemma, its morphological features) with aspects of a word's linguistic
context (e.g. whether it lies between two entities or on a dependency
path between them). While these help learning, they make generalization to unseen words
difficult. An alternative approach to capturing lexical information relies on 
\emph{continuous} word embeddings\footnote{Such embeddings
  have a long history in NLP, including term-document frequency
  matrices and their low-dimensional counterparts obtained by linear
  algebra tools (LSA, PCA, CCA, NNMF), Brown clusters, random
  projections and vector space models. Recently, neural networks / deep
learning have provided several popular methods for obtaining such
embeddings.} as representative of words but generalizable to new words.
Embedding features have improved many tasks,
including NER, chunking, dependency parsing, semantic role
labeling, and relation extraction
\cite{miller_name_2004,turian2010word,koo_simple_2008,roth_composition_2014,sun_semi-supervised_2011,plank_embedding_2013,nguyen_employing_2014}.
Embeddings can capture lexical information, but alone they are
insufficient: in state-of-the-art systems, they are used
alongside features of the broader linguistic context.

In this paper, we introduce a compositional model that combines unlexicalized
linguistic context and word embeddings for relation extraction, a task in which contextual feature
construction plays a major role in generalizing to unseen data.
Our model allows for the composition of embeddings with arbitrary linguistic structure, as
expressed by hand crafted features.
In the following sections, we begin with a
precise construction of compositional embeddings using word embeddings
in conjunction with unlexicalized features. Various feature
sets used in prior work
\cite{turian2010word,nguyen_employing_2014,hermann-EtAl:2014:P14-1,roth_composition_2014}
are captured as special cases of this construction.
Adding these
compositional embeddings directly to a standard log-linear model
yields a special case of our full model.
\begin{savenotes}
\begin{table*}[htbp]
\centering
\small
\begin{tabular}{|p{.03cm}l|c|c|c|}
\hline
& \bf Class & \bf M$_1$ & \bf M$_2$ & \bf Sentence Snippet\\
\hline
(1) & ART(M$_1$,M$_2$) & a man & a taxicab & \textit{A  man \colorA{driving} what
appeared  to be a taxicab} \\
(2) & PART-WHOLE(M$_1$,M$_2$) &  the  southern  suburbs & Baghdad &\textit{
direction  of  the  southern  suburbs \colorB{of}  Baghdad}\\
(3) & PHYSICAL(M$_2$,M$_1$) &  the united states &284 people & \textit{in the
united states , 284 people \colorC{died}} \\
 \hline
\end{tabular}
\vspace{-0.05 in}
\caption{\small{Examples from ACE 2005.
In (1) the word
``\colorA{driving}'' is a strong indicator of the relation \textit{ART}\footnote{In ACE 2005, \textit{ART} refers to a relation between a person and an artifact; such as a user, owner, inventor, or manufacturer relationship}
between
$M_1$ and $M_2$. A feature that depends on the embedding for this
context word could generalize to other lexical
indicators of the same relation (e.g. ``operating'') that don't appear
with \textit{ART} during training.
But lexical information alone is insufficient; relation extraction requires the identification of lexical roles: \textit{where} a word appears
structurally in the sentence. In (2), the word ``\colorB{of}'' between
``suburbs'' and ``Baghdad'' suggests that the first entity is part of the second, yet
the earlier occurrence after ``direction'' is of no significance to
the relation. 
Even finer information can be expressed by a word's role on
the dependency path between entities. In (3) we can distinguish the word
``\colorC{died}'' from other irrelevant words that don't appear between
the entities.}
}
\label{tab:example}
\vspace{-0.15 in}
\end{table*}
\end{savenotes}
We then treat the word embeddings as parameters giving rise
to our powerful, efficient, and easy-to-implement \emph{log-bilinear
model}. The model capitalizes on arbitrary types of
linguistic annotations by better utilizing features associated with
substructures of those annotations, including global information. We
choose features to promote different properties and to distinguish
different functions of the input words.

The full model involves three stages. First, it 
decomposes the annotated sentence into substructures
(i.e. a word and associated annotations). Second, it extracts features for each substructure (word),
and combines them with the word's embedding to
form a \emph{substructure embedding}.  
Third, we sum over substructure embeddings to
form a composed \emph{annotated sentence embedding},
which is used by a final softmax layer
to predict the output label (relation).

The result is a state-of-the-art relation extractor
for unseen domains from ACE 2005 \cite{walker_ace_2006}
and the relation classification dataset from SemEval-2010 Task 8 \cite{Hendrickx:EtAl:10}.

\paragraph{Contributions}

This paper makes several contributions, including:
\begin{compactenum}
\item We introduce the \fct{}, a new compositional embedding model for
  relation extraction.
\item We obtain the best reported results on ACE-2005 for
  coarse-grained relation extraction in the cross-domain setting, by
  combining \fct{} with a log-linear model.
\item We obtain results on on SemEval-2010 Task 8 competitive
  with the best reported results.
\end{compactenum}
\noindent Note that other work has already been published that builds 
on the \fct{}, such as \newcite{hashimoto2015task},
\newcite{nguyenrelation},
\newcite{santos2015classifying}, \newcite{yu2015learning} and \newcite{yu_combining_2015}.
Additionally, we have extended \fct{}
to incorporate a low-rank embedding of the features \cite{yu_combining_2015}, which
focuses on fine-grained relation extraction for ACE and
ERE.
This paper
obtains better results than the low-rank extension on
ACE coarse-grained relation extraction.


\section{Relation Extraction}
\label{sec:background}

In relation extraction we are given a sentence as input with the goal of identifying, for all pairs of entity mentions,
what relation exists between them, if any. For each pair of entity mentions in a sentence $S$,
we construct an instance 
$(y, \bx)$, where $\bx = (M_1, M_2, S, A)$.
$S=\{w_1,w_2,...,w_n\}$ is a sentence of length $n$ that expresses
a relation of type $y$ between two entity mentions $M_1$ and $M_2$, where
$M_1$ and $M_2$ are sequences of words in $S$. $A$ is
the associated annotations of sentence $S$, such as part-of-speech
tags, a dependency parse, and named entities. 
We consider directed relations: for a relation type $Rel$,
$y$$=$$Rel(M_1,M_2)$ and $y'$$=$$Rel(M_2,M_1)$ are different relations.
\tabref{tab:example} shows ACE 2005 relations, and has a strong label bias
towards negative examples. We also consider the task of relation 
\emph{classification} (SemEval), where the number
of negative examples is artificially reduced. 

\paragraph{Embedding Models}
Word embeddings and compositional embedding models have 
been successfully applied to a range of NLP tasks, 
however the applications of these embedding models 
to relation extraction are still limited.
Prior work on relation classification (e.g. SemEval 2010
Task 8) has focused on short sentences with at most one relation per sentence
\cite{socher-EtAl:2012:EMNLP-CoNLL,zeng-EtAl:2014:Coling}.
For relation extraction, where negative examples abound, prior work
has assumed that only the
named entity boundaries and not their types were available
\cite{plank_embedding_2013,nguyen-plank-grishman:2015:ACL-IJCNLP}. Other work 
has assumed that the order of two entities in a relation are given while
the relation type itself is unknown
\cite{nguyen_employing_2014,nguyenrelation}.
%
%
The standard relation extraction task, as adopted by
ACE 2005 \cite{walker_ace_2006}, uses long sentences containing
multiple named entities with known types\footnote{Since the focus of this paper is
relation extraction, we adopt the evaluation setting of prior work
which uses gold named entities to better facilitate comparison. }
and unknown relation directions.
We are the first to apply neural language model embeddings to this
task.



\paragraph{Motivation and Examples}
\label{ssec:emb_fea}

Whether a word is indicative of a relation depends on multiple properties,
which may relate to its context within the sentence.
For example, whether the word is in-between the entities, on the
dependency path between them, or to their left or right may provide
additional complementary information. Illustrative examples are
given in \tabref{tab:example} and provide the motivation for our
model.
In the next section, we will show how we develop informative representations
capturing both the semantic information in word embeddings and the contextual
information expressing a word's role relative to the entity mentions.
We are the first to incorporate all of this information at once. The closest work is that
of \newcite{nguyen_employing_2014}, who use a log-linear model for
relation extraction with embeddings as features for only the entity heads.
Such embedding features are insensitive to the broader contextual
information and, as we show, are not sufficient to elicit the word's role in a relation.

\section{A Feature-rich Compositional Embedding Model for Relations}
\label{sec:fct}

We propose a general framework to construct an embedding
of a sentence with annotations on its component words.
While we focus on the relation extraction task, the framework applies to any task
that benefits from both embeddings and typical hand-engineered lexical
features.
%


\begin{figure*}[tbp]
\centering
\includegraphics[scale=0.5]{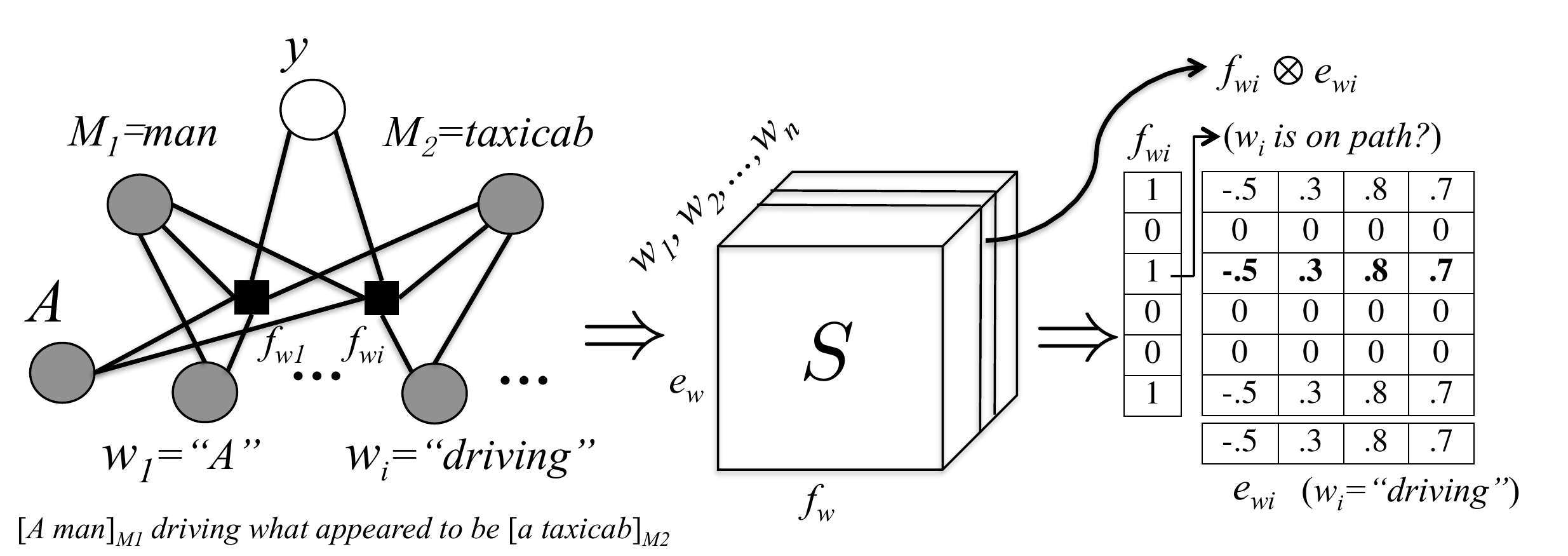}
\caption{\small{ Example construction of substructure embeddings.
Each substructure is a word $w_i$ in $S$, augmented by the
  target entity information and related information from annotation
  $A$ (e.g. a dependency tree).  We show the
  factorization of the annotated sentence into substructures (left),
  the concatenation of the substructure embeddings for the sentence
  (middle), and a single substructure embedding from that
  concatenation (right). The annotated sentence embedding (not shown)
  would be the sum of the substructure embeddings, as opposed to their
  concatenation.  }}
\label{fig:input}
\vspace{-0.1 in}
\end{figure*}

\subsection{Combining Features with Embeddings}
We begin by describing a precise method for
constructing substructure embeddings and annotated sentence embeddings
from existing (usually unlexicalized) features and embeddings. 
Note
that these embeddings can be included directly in a log-linear
model as features---doing so results in a special case of our full model presented
in the next subsection.

An annotated sentence is first decomposed into substructures. The
type of substructures can vary by task; for relation extraction we
consider one substructure per word\footnote{We use words as substructures
for relation extraction, but use the general terminology to maintain model generality.}.
For each substructure in the
sentence we have a hand-crafted feature
vector $f_{w_i}$ and a dense embedding vector $e_{w_i}$. 
We represent each substructure as the outer product $\otimes$
between these two vectors to
produce a matrix, herein called a \textbf{substructure embedding}: 
$  h_{w_i} = f_{w_i} \otimes e_{w_i}  $.
The features $f_{w_i}$ are based on the local context in $S$ and
annotations in $A$, which can include global information about the
annotated sentence.  These features allow the model to promote
different properties and to distinguish different functions of the
words.  Feature engineering can be task specific, as relevant
annotations can change with regards to each task. In this work we
utilize unlexicalized binary features common in relation extraction.
Figure \ref{fig:input} depicts the construction of a sentence's substructure
embeddings. 

We further sum over the substructure embeddings to
form an \textbf{annotated sentence embedding}:
\begin{align}
e_{x} = \sum_{i=1}^n  f_{w_i} \otimes e_{w_i}
\end{align}
When both the hand-crafted features and word embeddings are treated as
inputs, as has previously been the case in relation extraction, this
annotated sentence embedding can be used directly as the features of a
log-linear model. In fact, we find that the feature sets used in prior
work for many other NLP tasks are special cases of this simple
construction
\cite{turian2010word,nguyen_employing_2014,hermann-EtAl:2014:P14-1,roth_composition_2014}. This
highlights an important connection: when the word embeddings are
constant, our constructions of substructure and annotated sentence
embeddings are just specific forms of polynomial (specifically
quadratic) feature combination---hence their commonality in the literature.
Our experimental results suggest that such a construction is more
powerful than directly including embeddings into the model.

\subsection{The Log-Bilinear Model}
\label{ssec:model}

Our full log-bilinear model 
first forms the substructure and annotated sentence embeddings from
the previous subsection. The model uses its parameters to score
the annotated sentence  embedding and uses a softmax to produce an output label.
We call the entire model the \textbf{Feature-rich Compositional Embedding Model (\fct)}.

Our task is to determine the label $y$ (relation) given the instance $\bx = (M_1, M_2, S,
A)$. We formulate this as a probability.
\begin{equation}
P(y | \bx; T, \mathbf{e}) = \frac{ \exp\left( \sum_{i=1}^n T_y \odot \left( f_{w_i} \otimes e_{w_i} \right) \right)  }{ Z(\bx) }
\label{eq:fct}
\end{equation}
where $\odot$ is
the `matrix dot product' or Frobenious inner product of the two
matrices. The normalizing constant which sums over all possible output
labels $y' \in L$ is given by $Z(\bx) = \sum_{y' \in L} \exp\left(
  \sum_{i=1}^n T_{y'} \odot \left( f_{w_i} \otimes e_{w_i} \right)
\right)$.
The parameters of the model are the word embeddings $\mathbf{e}$ for
each word type and a list of weight matrix $T=[T_y]_{y\in L}$ which is used to score each label
$y$. The model is log-bilinear
\footnote{Other popular log-bilinear
models are the log-bilinear language models \cite{mnih_three_2007,mikolov2013distributed}.} 
(i.e. log-quadratic) since we recover a
log-\emph{linear} model by fixing either
$\mathbf{e}$ or $T$. We study
both the full log-bilinear and the log-linear model obtained by
fixing the word embeddings.



\subsection{Discussion of the Model}

\paragraph{Substructure Embeddings}
Similar words (i.e. those with similar embeddings) with
similar functions in the sentence (i.e. those with similar features)
will have similar matrix representations.
To understand our selection of the outer product, consider the example in Fig. \ref{fig:input}.
The word ``driving'' can indicate the \textit{ART} relation if it appears on
the dependency path between $M_1$ and $M_2$.
Suppose the third feature in $f_{w_i}$ indicates this \featnormal{on-path} feature.
Our model can now 
learn parameters which give the third row a high weight for the \textit{ART} label.
Other words with embeddings similar to ``driving'' that appear on the dependency path
between the mentions will similarly receive high weight for the \textit{ART} label.
On the other hand, if the embedding is similar but is not on the dependency path, it will have 0 weight.
Thus, our model generalizes its model parameters across words with similar embeddings
only when they share similar functions in the sentence.

\paragraph{Smoothed Lexical Features}
Another intuition about the selection of outer product is that it is actually a smoothed version
of traditional lexical features used in classical NLP systems.
Consider a lexical feature $f=u\wedge w$, which is a conjunction (logic-and) between 
non-lexical property $u$ and lexical part (word) $w$. 
If we represent $w$ as a one-hot vector, then the outer product 
exactly recovers the original feature $f$.
Then if we replace the one-hot representation with its word embedding, 
we get the current form of our \fct. 
Therefore, our model can be viewed as a smoothed version of lexical features, which keeps the expressive strength, and uses 
embeddings to generalize to low frequency features.

\paragraph{Time Complexity}
Inference in \fct\ is much faster than both CNNs \cite{collobert2011natural} and RNNs
\cite{socher-EtAl:2013:EMNLP,bordes2012semantic}.
\fct\  requires $O(snd)$ products on average with 
sparse features, where $s$ is the average number of per-word non-zero
feature values, $n$ is the length of the sentence, and
$d$ is the dimension of word embedding. In contrast, CNNs and RNNs
usually have complexity $O(C \cdot n d^2)$, where $C$ is a
model dependent constant.

\section{Hybrid Model}
\label{sec:hybrid}

We present a hybrid model which combines the \fct{}
with an existing log-linear model. We do so by defining a new model:
\begin{align}
p_{\text{\fct{}+loglin}}(y | x) = \frac{1}{Z} p_{\fct{}}(y|x) p_{\text{loglin}}(y|x)
  \label{eq:hybrid}  
\end{align}
The log-linear model has the usual form: $p_{\text{loglin}}(y|x)
\propto \exp(\vc{\theta} \cdot \vc{f}(x,y))$, where
$\vc{\theta}$ are the model parameters and $\vc{f}(x,y)$ is a vector of
features. The integration treats each model as a providing a score
which we multiply together. The constant $Z$ ensures a
 normalized distribution.

\section{Training}

\fct\ training optimizes a
cross-entropy objective:
\[
\ell(D;T,\mathbf{e})  = \sum_{(\bx, y)\in D} \log P(y |
\bx; T,\mathbf{e})
\]
\noindent where $D$ is the set of all training data and $\mathbf{e}$ is the set
of word embeddings. To optimize the objective, 
for each instance $(y, \bx)$ we perform stochastic
training on the
loss function $\ell = \ell(y, \bx; T,\mathbf{e}) = \log
P(y | \bx; T, \mathbf{e})$. 
The gradients of the model parameters are obtained by backpropagation
(i.e. repeated application of the chain rule). 
We define the vector $\mathbf s=\left[ \sum_i T_y
\odot (f_{w_i} \otimes e_{w_i}) \right]_{1\le y \le L}$, 
which yields 
\begin{align}
\frac{\partial \ell}{\partial \mathbf s} =\left[ \left(
    I[y'=y]-P(y' | \bx;T,\mathbf{e}) \right)_{1 \le y \le L}
\right]^T, \nonumber
\end{align}
where the indicator function $I[x]$ equals 1 if $x$
is true and 0 otherwise.
We have the following gradients:
$\frac{\partial \ell}{\partial T} = \frac{\partial \ell}{\partial
\mathbf s} \otimes \sum_{i=1}^n   {f_{w_i}} \otimes  e_{w_i}$,
which is equivalent to:
\begin{align}
\frac{\partial \ell}{\partial T_{y'}} = \left( I[y=y']-P(y' | \bx; T,\mathbf{e}) \right) \cdot 
\sum_{i=1}^n  f_{w_i} \otimes e_{w_i}. \nonumber
\end{align}

\noindent When we treat the word embeddings as parameters (i.e. the log-bilinear model),
we also fine-tune the word embeddings with the \fct\ model:
\begin{align}
\frac{\partial \ell}{\partial e_w} 
 &= \sum_{i=1}^n \left(\left( \sum_y \frac{\partial \ell}{\partial \mathbf s_y} T_y\right) \cdot {f_i}  \right) \cdot  I[w_i= w] . \nonumber
\end{align}

As is common in deep learning, we
initialize these embeddings from an neural
language model and then \emph{fine-tune} them for our supervised task.
The training process for the hybrid model (\secref{sec:hybrid}) is
also easily done by backpropagation since each sub-model has separate
parameters.

\section{Experimental Settings}
\label{sec:exp_setting}

\paragraph{Features}

Our \fct{} features (\tabref{tab:fea}) use a feature vector $f_{w_i}$ over the word $w_i$, the two target
entities $M_1, M_2$, and their dependency path.
Here $h_1,h_2$ are the indices of the two head words of
$M_1,M_2$,
$\times$ refers to the Cartesian product between two sets, $t_{h_1}$ and
$t_{h_2}$ are entity types (named entity tags for ACE 2005 or WordNet supertags for SemEval 2010) of the head
words of two entities, and $\phi$ stands for the empty feature.
$\oplus$ refers to the conjunction of two elements.
The \featnormal{In-between} features indicate whether a word $w_i$ is in
between two target entities, and the \featnormal{On-path} features indicate
whether the word is on the dependency path, on which there is a set of words $P$, between the two entities. 


We also use the target entity type as a feature.
Combining this with the basic features results in more powerful compound features,
which can help us better 
distinguish the functions of word embeddings for predicting certain relations. For example, if 
we have a person and a vehicle, we know it will be more likely that they have an 
\textit{ART} relation. For the \textit{ART} relation, we introduce a corresponding 
weight vector, which is closer to lexical embeddings similar to the embedding of ``drive''.

All linguistic annotations needed for features (POS, chunks\footnote{Obtained from the constituency parse
using the CONLL 2000 chunking converter (Perl script).}, parses) are from Stanford CoreNLP \cite{manning-EtAl:2014:P14-5}. 
Since SemEval does not have gold entity types we
obtained WordNet  and named entity tags using 
\newcite{ciaramita-altun:2006:EMNLP}.
For all experiments we use 200-d word embeddings trained on the NYT
portion of the Gigaword 5.0 corpus \cite{parker2011english}, with word2vec 
\cite{mikolov2013distributed}. We use the CBOW model
with negative sampling (15 negative words). We set a window size $c$=5, and 
remove types occurring less
than 5 times.

\begin{table}[tb]
\centering
\scriptsize
\begin{tabular}{|l|c|}
\hline
\bf Set & \bf Template\\
\hline
\feat{HeadEmb} & $\{I[i=h_1], I[i=h_2]\}$ \\
& ($w_i$ is head of $M_1/M_2$) $\times \{\phi, t_{h_1},t_{h_2},t_{h_1}\oplus t_{h_2}\}$\\
\hline
\feat{Context} &  $I[i=h_1\pm 1]$ (left/right token of $w_{h_1}$) \\
&  $I[i=h_2\pm 1]$ (left/right token of $w_{h_2}$) \\
\hline
 \feat{In-between}  & $I[i > h_1] \& I[i < h_2]$ (in between ) \\
 & $\times \{\phi, t_{h_1},t_{h_2},t_{h_1}\oplus t_{h_2}\}$\\
 \hline
 \feat{On-path} & $I[w_i \in P]$ (on path)  \\
 & $\times \{\phi, t_{h_1},t_{h_2},t_{h_1}\oplus t_{h_2}\}$ \\
 \hline
\end{tabular}
\vspace{-0.1 in}
\caption{Feature sets used in \fct.}
\label{tab:fea}
\vspace{-0.1 in}
\end{table}


\paragraph{Models}

We consider several methods. 
(1) \fct{} in isolation without fine-tuning.  
(2) \fct{} in isolation with fine-tuning (i.e. trained as a log-bilinear model).
(3) A log-linear model 
with a rich binary feature set from
\newcite{sun_semi-supervised_2011} (Baseline)---this consists of all
the baseline features of \newcite{zhou_exploring_2005} plus several additional
carefully-chosen features that have been highly tuned for ACE-style
relation extraction over years of research. We exclude the Country
gazetteer and WordNet features from \newcite{zhou_exploring_2005}.
The two remaining methods are hybrid models that integrate \fct{} as a submodel within the
log-linear model (\secref{sec:hybrid}). We consider two 
combinations.
(4) The feature set of \newcite{nguyen_employing_2014} obtained by using the embeddings of
heads of two entity mentions (+HeadOnly). 
(5) Our full \fct\ model (+\fct).
All models use L2 regularization tuned on dev data.

\subsection{Datasets and Evaluation}

\paragraph{ACE 2005}
We evaluate our relation extraction system on the English portion of the
ACE 2005 corpus \cite{walker_ace_2006}.\footnote{Many relation extraction systems
  evaluate on the ACE 2004 corpus \cite{mitchell_ace_2005}. Unfortunately, the
  most common convention is to use 5-fold cross validation,
  treating the entirety of the dataset as both train \emph{and}
  evaluation data. 
  Rather than continuing to overfit this data by perpetuating
  the cross-validation convention, we instead focus on ACE 2005.}
There are 6 domains: Newswire (\texttt{nw}), Broadcast
Conversation (\texttt{bc}), Broadcast News (\texttt{bn}), Telephone
Speech (\texttt{cts}), Usenet Newsgroups (\texttt{un}), and Weblogs
(\texttt{wl}). Following prior work
we focus on the domain adaptation
setting, where we train on one set
(the
union of the news domains (\texttt{bn}+\texttt{nw}),
tune hyperparameters on a dev domain (half of
\texttt{bc}) and evaluate on the remainder (\texttt{cts}, \texttt{wl}, and the remainder of
\texttt{bc}) \cite{plank_embedding_2013,nguyen_employing_2014}.
%
%
We assume that 
 gold entity spans
\emph{and} types are available for train and test. We use all
pairs of entity mentions to yield 43,518 total relations in the
training set. 
%
%
We report precision, recall, and F1 for relation
extraction.
While it is not our focus, for completeness we include results with
unknown entity types following \newcite{plank_embedding_2013} (Appendix 1).

\paragraph{SemEval 2010 Task 8}
We evaluate on the SemEval 2010 Task 8
dataset\footnote{\tiny{\url{http://docs.google.com/View?docid=dfvxd49s_36c28v9pmw}}} \cite{Hendrickx:EtAl:10}
to compare with other compositional models and highlight the advantages of \fct{}.
This task is to determine
the relation type (or no relation) between two entities in a
sentence.
We adopt the setting of
\newcite{socher-EtAl:2012:EMNLP-CoNLL}. 
%
We use 10-fold cross validation on the training data to select hyperparameters
and do regularization by early stopping.
The learning rates for \fct\ with/without
fine-tuning are 5e-3 and 5e-2 respectively.  We report macro-F1 and
compare to previously published results.

\section{Results}
\label{sec:exp}

\paragraph{ACE 2005}
\label{ssec:res_ace}

\begin{table*}[t]
\centering
\small
\begin{tabular}{|p{.1cm}l|c|c|c|c|c|c|c|c|c|c|}
\hline
& \multirow{2}{*} & \multicolumn{3}{|c|}{\bf bc} &
 \multicolumn{3}{|c|}{\bf cts} & \multicolumn{3}{|c|}{\bf wl} & \bf Avg.\\
\cline{3-11}
& \bf Model & \bf P & \bf R & \bf F1& \bf P & \bf R & \bf F1& \bf P &
\bf R & \bf F1 & \bf  F1\\
        \hline
      (1) & \fct\ only (ST) & 66.56 & \textbf{57.86} & 61.90 & 65.62 & 44.35 & 52.93 & 57.80 & 44.62 & 50.36 & 55.06 \\
      (3) & Baseline (ST) & \textbf{74.89} & 48.54 & 58.90 & 74.32 & 40.26 & 52.23 & 63.41 & 43.20 & 51.39 & 54.17 \\
      (4) & + HeadOnly (ST)& 70.87 & 50.76 & 59.16 & 71.16 & 43.21 & 53.77 & 57.71 & 42.92 & 49.23 & 54.05 \\
       (5) & + \fct\ (ST) & 74.39 & 55.35 & \textbf{63.48} & \textbf{74.53} & \textbf{45.01} & \textbf{56.12} & \textbf{65.63} & \textbf{47.59} & \textbf{55.17} & \textbf{58.26} \\
        \hline
\end{tabular}
\caption{Comparison of models on ACE 2005 out-of-domain test sets. 
%
%
Baseline + HeadOnly is our reimplementation
of the features of
\newcite{nguyen_employing_2014}.
}
\label{tab:tmp_sub_results}
\end{table*}

%


Despite \fct{}'s (1) simple feature set, it is competitive with the log-linear
baseline (3) on out-of-domain test sets (\tabref{tab:tmp_sub_results}).
%
In the typical gold entity spans and types setting, both
\newcite{plank_embedding_2013} and 
\newcite{nguyen_employing_2014} found that they were unable to obtain improvements
by adding embeddings to baseline feature sets. By contrast, we
find that on all domains the combination baseline + \fct\ (5) obtains the highest F1
and significantly outperforms the other baselines, {\em yielding the best reported
results for this task.}
We found that fine-tuning of embeddings (2) did not yield improvements on our
out-of-domain development set, in contrast to our results below for 
SemEval. We suspect this is
because fine-tuning allows the model to overfit the training domain,
which then hurts performance on the unseen ACE test domains.
Accordingly, \tabref{tab:tmp_sub_results} shows only
the log-linear model.

Finally, we highlight an important contrast between \fct{} (1) and
the log-linear model (3): the latter uses over 50
feature templates based on a POS tagger, dependency parser,
chunker, and constituency parser. \fct\ uses only a 
dependency parse
but still obtains better results (Avg. F1).

\begin{table*}[tb]
\centering
\small
\begin{tabular}{|l|l|c|}
\hline
\bf Classifier & \bf Features & \bf F1\\
\hline
SVM \cite{rink-harabagiu:2010:SemEval}  &POS, prefixes, morphological, WordNet, dependency parse,
 & \multirow{3}{*}{82.2} \\
(Best in SemEval2010)	&Levin classed, ProBank, FrameNet, NomLex-Plus, &\\
&Google n-gram, paraphrases, TextRunner&\\
\hline
RNN &  word embedding, syntactic parse & 74.8\\
RNN + linear&  word embedding, syntactic parse, POS, NER, WordNet & 77.6 \\
\hline
 MVRNN  & word embedding, syntactic parse & 79.1\\
 MVRNN + linear& word embedding, syntactic parse, POS, NER, WordNet & 82.4\\
 \hline
 CNN \cite{zeng-EtAl:2014:Coling}  & word embedding, WordNet & 82.7\\
 \hline
 CR-CNN (log-loss) & word embedding & 82.7\\ 
 CR-CNN (ranking-loss) & word embedding & \textbf{84.1}\\
 \hline
  RelEmb (word2vec embedding) & word embedding & 81.8\\
 RelEmb (task-spec embedding) & word embedding & 82.8\\
  RelEmb (task-spec embedding)  + linear & word embedding, dependency paths, WordNet, NE & 83.5\\
 \hline
  DepNN & word embedding, dependency paths& 82.8\\
  DepNN  + linear & word embedding, dependency paths, WordNet, NER & 83.6\\
 \hline
 \multirow{2}{*}{(1) \fct\ (log-linear)} & word embedding, dependency parse, WordNet & 82.0\\
 & word embedding, dependency parse, NER & 81.4\\
 \hline
 \multirow{2}{*}{(2) \fct\ (log-bilinear)} & word embedding, dependency parse, WordNet &  {82.5}\\
 & word embedding, dependency parse, NER & 83.0\\
 \hline
 \multirow{2}{*}{(5) \fct\ (log-linear) + linear (Hybrid)} & word embedding, dependency parse, WordNet & 83.1\\
 & word embedding, dependency parse, NER & \textbf{83.4}\\
 \hline
\end{tabular}
\caption{Comparison of \fct{} with previously published results for SemEval 2010 Task 8.}
\label{tab:res}
\end{table*}

\paragraph{SemEval 2010 Task 8}
\label{ssec:res_semeval}

Table \ref{tab:res} shows \fct\ compared to the best reported results
from the SemEval-2010 Task 8 shared task and several other 
compositional models. 

For the \fct\ we considered two feature sets. We found that using NE
tags instead of WordNet tags helps with fine-tuning but hurts
without. This may be because the set of WordNet tags is larger making
the model more expressive, but also introduces more parameters.  When
the embeddings are fixed, they can help to better distinguish
different functions of embeddings. But when fine-tuning, it becomes
easier to over-fit. Alleviating over-fitting is a subject for future
work (\secref{sec:conclusion}).

With either WordNet or NER features, \fct{} achieves better
performance than the RNN and MVRNN. With NER features and fine-tuning,
it outperforms a CNN \cite{zeng-EtAl:2014:Coling} and also the
combination of an embedding model and a traditional log-linear model
(RNN/MVRNN + linear) \cite{socher-EtAl:2012:EMNLP-CoNLL}.
As with ACE, \fct{} uses less linguistic resources than many close competitors \cite{rink-harabagiu:2010:SemEval}.

We also compared to concurrent work on enhancing the compositional models with task-specific 
information for relation classification, including
\newcite{hashimoto2015task} (RelEmb), which trained task-specific word embeddings,
and \newcite{santos2015classifying} (CR-CNN), which proposed a task-specific ranking-based loss function.
Our Hybrid methods (\fct{} + linear) get comparable results to theirs. 
Note that their base compositional model results without any task-specific enhancements,
i.e. RelEmb with word2vec embeddings and CR-CNN with log-loss, are still lower than
the best \fct{} result. We believe that \fct{} can be also improved with
these task-specific enhancements, e.g. replacing the word embeddings to the task-specific
ones from \cite{hashimoto2015task} increases the result to 83.7\% (see \S\ref{ssec:emb_res} for details).
We leave the application of ranking-based loss to future work.

Finally, a concurrent work 
\cite{liu-EtAl:2015:ACL-IJCNLP} proposes DepNN, which builds representations for the 
dependency path (and its attached subtrees) between two entities by applying recursive
and convolutional neural networks successively.
Compared to their model, our \fct{} achieves comparable results.
Of note, our \fct{} and the RelEmb are also the most efficient models among all above
compositional models since they have linear time complexity with respect to the dimension of 
embeddings.


\subsection{Effects of the embedding sub-models}
We next investigate the effects of different types
of features on \fct{} using ablation tests on ACE 2005 (\tabref{tab:fct_sub_results}.) We focus on \fct{} alone with the feature templates of \tabref{tab:fea}.
Additionally, we show results of using \emph{only} the head embedding features from
\newcite{nguyen_employing_2014} (HeadOnly).
Not surprisingly, the HeadOnly model performs poorly (F1 score = 14.30\%),
showing the importance of our rich binary feature set.
Among all the features templates, removing \featnormal{HeadEmb} results in the
largest degradation.
The second most important feature template is \featnormal{In-between}, while
\featnormal{Context} features have little impact.
Removing all entity type features ($t_{h_i}$) does 
significantly worse than the full model,
showing the value of our entity type features.
\SideNoteMD{Someone in the above section we need to say that we now
  have the best results on ACE.}
\SideNoteMRG{For now, we're sticking with `we attain state-of-the-art'
  in the intro. Nothing stronger.}
\SideNoteMY{details(MD)}

\begin{table}[tbp]
\centering
\small
\begin{tabular}{|l|c|c|c|}
\hline
\bf Feature Set & \bf Prec & \bf Rec & \bf F1\\
\hline
\multirow{1}{*}{HeadOnly} &31.67 & 9.24 & 14.30\\
\hdashline
 \fct\ & 69.17 & \bf 56.73 & \bf 62.33\\
\quad-\featnormal{HeadEmb} & 66.06 & 47.00 & 54.92 \\
 \quad-\featnormal{Context}  & 70.89 & 55.27 & 62.11\\
 \quad-\featnormal{In-between}  & 66.39 & 51.86 & 58.23 \\
 \quad-\featnormal{On-path} &  69.23 & 53.97 & 60.66 \\
 \hdashline
 \fct-\featnormal{EntityTypes} & \bf 71.33 & 34.68 & 46.67\\
 \hline
\end{tabular}
\caption{Ablation test of \fct\ on development set.}
\label{tab:fct_sub_results}
\end{table}

\subsection{Effects of the word embeddings}
\label{ssec:emb_res}
Good word embeddings are critical for both \fct{} and other compositional models. 
In this section, we show the results of \fct{} with embeddings used to
initialize other recent state-of-the-art models. 
Those embeddings include the 300-$d$ baseline embeddings trained on English Wikipedia (w2v-enwiki-d300) and the 100-$d$ task-specific embeddings (task-specific-d100)\footnote{
In the task-specific setting, \fct{} will represent entity words and context words with separate sets of embeddings.}
from the RelEmb paper \cite{hashimoto2015task}, the 400-$d$ embeddings from the CR-CNN paper \cite{santos2015classifying}.
Moreover, we list the best result (DepNN) in \newcite{liu-EtAl:2015:ACL-IJCNLP}, which uses the same
embeddings as ours.
\tabref{tab:emb_res} shows the effects of word embeddings on \fct{}
and provides relative comparisons between \fct{} and the other state-of-the-art models.
We use the same hyperparameters and number of iterations in
\tabref{tab:res}.

The results show that using different embeddings to initialize \fct{}
can improve F1 beyond our previous results. We also find that
increasing the dimension of the word embeddings does not necessarily lead to better results due to the
problem of over-fitting (e.g.w2v-enwiki-d400 vs. w2v-enwiki-d300).
With the same initial embeddings, \fct{} usually gets better results
without any changes to the hyperparameters than the competing model,
further confirming the advantage of \fct{} at the model-level as discussed under \tabref{tab:res}.
The only exception is the DepNN model, which gets better result than \fct{} on the same embeddings.
The task-specific embeddings from \cite{hashimoto2015task} leads to the
best performance (an improvement of 0.7\%). This observation suggests that the other compositional models may also benefit from the work of \newcite{hashimoto2015task}.
 
\begin{table}[tb]
\centering
\small
\begin{tabular}{|l|l|c|}
\hline
\bf Embeddings & \bf Model & \bf F1\\
\hline
\multirow{2}{*}{w2v-enwiki-d300} &  RelEmb  & 81.8\\
 & {(2) \fct\ (log-bilinear)} & 83.4\\
\hline
\multirow{3}{*}{task-specific-d100}  & RelEmb & 82.8\\
& RelEmb+linear & 83.5\\
& {(2) \fct\ (log-bilinear)} & \bf 83.7\\
 \hline
 \multirow{2}{*}{w2v-enwiki-d400} & CR-CNN & 82.7\\
 &{(2) \fct\ (log-bilinear)} & 83.0\\
 \hline
 \multirow{2}{*}{w2v-nyt-d200} & DepNN   & 83.6\\
 & {(2) \fct\ (log-bilinear)}  & 83.0\\
 \hline
\end{tabular}
\caption{Evaluation of \fct{}s with different word embeddings on SemEval 2010 Task 8.}
\label{tab:emb_res}
\end{table}

\section{Related Work}
\label{sec:related}

\NoteMY{Move the discussion about disadvantages of Feature Engineering early: Lexicalized features are often hand designed to capture the linguistic
interactions between words. Yet, such features don't generalize to
new words not seen during training nor do they allow sharing of statistical strength between
lexically similar substructures: for example, learning that the bigram
`father of' intervening between two entity mentions is indicative of a
familial relation won't allow for generalization to the bigram `mother
of' which expresses a similar relation. Back-off features are often employed using the word's prefix,
suffix, lemma, or POS tag---but these may miss the semantic
connections required for the previous example.
For example, back-off based on POS tags cannot distinguish the different roles of ``mother of" and ``director of" on relation extraction.
Word embeddings can
assist with exactly this issue.}
\NoteMY{move forward
Lexical features, although powerful, are easy to suffer from data sparsity.
In this situation, word embeddings, which represents word in some low-dimensional
space, become a important type of back-off features---in
part because they provide a simple method of semi-supervised learning.}
%

\NoteMY{
Usually, a major benefit to hand designing features is that one can
easily incorporate arbitrary substructures of the input.  However, the
methods of incorporation for dense, real-valued vectors have been
somewhat limited: commonly only a short list of word properties are
conjoined with a small set of word embeddings from the sentence (see
\S~\ref{sec:related} for further discussion of this point). Certainly,
one challenge along this direction is how to compose the vectors for
multiple words (e.g. the bigram to the left of a mention) into a
feature.}

\paragraph{Compositional Models for Sentences}
In order to build a representation
(embedding) for a sentence based on its component word
embeddings and structural information, 
recent work on compositional models (stemming from the deep learning
community) has designed model structures that mimic the structure of
the input. For example, 
these models could take into account the order of the
words (as in Convolutional Neural Networks (CNNs))
\cite{collobert2011natural} or build off of an
input tree (as in Recursive Neural Networks (RNNs) or the Semantic
Matching Energy Function)
\cite{socher-EtAl:2013:EMNLP,bordes2012semantic}.
\NoteMY{
Several results (c.f. \newcite{collobert2011natural}) have suggested
that this method of \emph{learning} features (as opposed to hand
designing them) is a promising avenue for supplanting traditional
feature engineering approaches. } 

While these models work well on
sentence-level representations, the nature of their designs
also limits them to fixed types of substructures from the annotated
sentence, such as chains for CNNs and trees for RNNs. Such models
cannot capture arbitrary combinations of linguistic annotations
available for a given task, such as word order, dependency tree, and
named entities used for relation extraction.
Moreover, these approaches ignore the differences in
functions between words appearing in different roles. 
This does not suit more general substructure labeling tasks in NLP, e.g. these models cannot be directly applied to relation extraction since they will output the same result for any pair of entities in a same sentence.


\paragraph{Compositional Models with Annotation Features}
To tackle the problem of traditional compositional models, \newcite{socher-EtAl:2012:EMNLP-CoNLL} made
the RNN model specific to relation extraction tasks by working on the minimal 
sub-tree which spans the two target entities. 
However, these
specializations to relation extraction does not generalize easily to
other tasks in NLP. There are two ways to achieve such specialization
in a more general fashion:

\textit{1. Enhancing Compositional Models with Features.}
A recent trend enhances compositional models with annotation features.
Such an approach has been shown to significantly improve
over pure compositional models.
For example,
\newcite{hermann-EtAl:2014:P14-1} and 
\newcite{nguyen_employing_2014}
gave different weights to words with different syntactic context types or
to entity head words with different argument IDs.
\newcite{zeng-EtAl:2014:Coling} use concatenations of embeddings as features
in a CNN model, according to their positions relative to the
target entity mentions.
\newcite{belinkov2014exploring} enrich embeddings with linguistic features 
before feeding them forward to a RNN model.
\newcite{socher2013parsing} and \newcite{hermann2013role} enhanced RNN models by refining
the transformation matrices with phrase types and CCG super tags.

\textit{2. Engineering of Embedding Features.}
A different approach to combining traditional linguistic features and embeddings is hand-engineering features with word 
embeddings and adding them to log-linear models.
Such approaches have achieved state-of-the-art results in
many tasks including NER, chunking, dependency parsing, 
semantic role labeling, and relation extraction
\cite{miller_name_2004,turian2010word,koo_simple_2008,roth_composition_2014,sun_semi-supervised_2011,plank_embedding_2013}.
\newcite{roth_composition_2014} considered features similar to ours
for semantic role labeling.

However, in prior work both of above approaches are only able to utilize limited information,
usually one property for each word.
Yet there may be different useful properties of a word which can contribute 
to the performances of the task. By contrast, our
\fct\ can easily utilize these features without changing the model
structures. 

In order to better utilize the dependency annotations,
recently work built their models according to the dependency paths
\cite{ma-EtAl:2015:ACL-IJCNLP,liu-EtAl:2015:ACL-IJCNLP}, 
which share similar motivations to the usage of \texttt{On-path} features in our work.

\paragraph{Task-Specific Enhancements for Relation Classification}

An orthogonal direction of improving compositional models for relation classification
is to enhance the models with task-specific information.
For example,
\newcite{hashimoto2015task} trained task-specific word embeddings,
and \newcite{santos2015classifying} proposed a ranking-based loss function for relation classification.




\section{Conclusion}
\label{sec:conclusion}
We have presented \fct, a new compositional model for deriving
sentence-level and substructure embeddings from word embeddings.  Compared to existing
compositional models, \fct\ can easily handle arbitrary types of input
and handle global information for composition, while remaining
easy to implement. We have demonstrated that  \fct\  alone attains near state-of-the-art
performances on several relation extraction tasks, and in 
combination with
traditional feature based log-linear models it 
obtains state-of-the-art results. 

Our next steps in improving \fct{} focus on enhancements
based on task-specific embeddings or loss functions as in 
\newcite{hashimoto2015task,santos2015classifying}.
Moreover, as the model provides
a general idea for representing both sentences and sub-structures in language, 
it has the potential to contribute useful components to various tasks,
such as dependency parsing, SRL and paraphrasing. 
Also as kindly pointed out by one anonymous reviewer,
our \fct\ can be applied to the TAC-KBP \cite{ji2010overview} tasks,
by replacing the training objective to a multi-instance multi-label one (e.g. \newcite{surdeanu2012multi}).
We plan to explore the above applications of \fct\  in the future. 

\section*{Acknowledgments}
We thank the anonymous reviewers for their comments, and Nicholas Andrews, Francis Ferraro, and Benjamin Van Durme for their input.
We thank Kazuma Hashimoto, C{\'i}cero Nogueira dos Santos, Bing Xiang and Bowen Zhou for sharing their word embeddings and many helpful discussions.
Mo Yu is supported by the China Scholarship Council and by NSFC 61173073.

\bibliographystyle{acl}
\bibliography{acere-2015.bib}

\begin{thebibliography}{}

\bibitem[\protect\citename{Belinkov \bgroup et al.\egroup
  }2014]{belinkov2014exploring}
Yonatan Belinkov, Tao Lei, Regina Barzilay, and Amir Globerson.
\newblock 2014.
\newblock Exploring compositional architectures and word vector representations
  for prepositional phrase attachment.
\newblock {\em Transactions of the Association for Computational Linguistics},
  2:561--572.

\bibitem[\protect\citename{Bordes \bgroup et al.\egroup
  }2012]{bordes2012semantic}
Antoine Bordes, Xavier Glorot, Jason Weston, and Yoshua Bengio.
\newblock 2012.
\newblock A semantic matching energy function for learning with
  multi-relational data.
\newblock {\em Machine Learning}, pages 1--27.

\bibitem[\protect\citename{Ciaramita and
  Altun}2006]{ciaramita-altun:2006:EMNLP}
Massimiliano Ciaramita and Yasemin Altun.
\newblock 2006.
\newblock Broad-coverage sense disambiguation and information extraction with a
  supersense sequence tagger.
\newblock In {\em EMNLP2006}, pages 594--602, July.

\bibitem[\protect\citename{Collobert \bgroup et al.\egroup
  }2011]{collobert2011natural}
Ronan Collobert, Jason Weston, L{\'e}on Bottou, Michael Karlen, Koray
  Kavukcuoglu, and Pavel Kuksa.
\newblock 2011.
\newblock Natural language processing (almost) from scratch.
\newblock {\em JMLR}, 12:2493--2537.

\bibitem[\protect\citename{dos Santos \bgroup et al.\egroup
  }2015]{santos2015classifying}
Cicero dos Santos, Bing Xiang, and Bowen Zhou.
\newblock 2015.
\newblock Classifying relations by ranking with convolutional neural networks.
\newblock In {\em Proceedings of the 53rd Annual Meeting of the Association for
  Computational Linguistics and the 7th International Joint Conference on
  Natural Language Processing (Volume 1: Long Papers)}, pages 626--634,
  Beijing, China, July. Association for Computational Linguistics.

\bibitem[\protect\citename{Hashimoto \bgroup et al.\egroup
  }2015]{hashimoto2015task}
Kazuma Hashimoto, Pontus Stenetorp, Makoto Miwa, and Yoshimasa Tsuruoka.
\newblock 2015.
\newblock Task-oriented learning of word embeddings for semantic relation
  classification.
\newblock {\em arXiv preprint arXiv:1503.00095}.

\bibitem[\protect\citename{Hendrickx \bgroup et al.\egroup
  }2010]{Hendrickx:EtAl:10}
Iris Hendrickx, Su~Nam Kim, Zornitsa Kozareva, Preslav Nakov, Diarmuid {\'O
  S\'eaghdha}, Sebastian Pad\'o, Marco Pennacchiotti, Lorenza Romano, and Stan
  Szpakowicz.
\newblock 2010.
\newblock Semeval-2010 task 8: Multi-way classification of semantic relations
  between pairs of nominals.
\newblock In {\em Proceedings of SemEval-2 Workshop}.

\bibitem[\protect\citename{Hermann and Blunsom}2013]{hermann2013role}
Karl~Moritz Hermann and Phil Blunsom.
\newblock 2013.
\newblock The role of syntax in vector space models of compositional semantics.
\newblock In {\em Association for Computational Linguistics}, pages 894--904.

\bibitem[\protect\citename{Hermann \bgroup et al.\egroup
  }2014]{hermann-EtAl:2014:P14-1}
Karl~Moritz Hermann, Dipanjan Das, Jason Weston, and Kuzman Ganchev.
\newblock 2014.
\newblock Semantic frame identification with distributed word representations.
\newblock In {\em Proceedings of the 52nd Annual Meeting of the Association for
  Computational Linguistics (Volume 1: Long Papers)}, pages 1448--1458,
  Baltimore, Maryland, June. Association for Computational Linguistics.

\bibitem[\protect\citename{Ji \bgroup et al.\egroup }2010]{ji2010overview}
Heng Ji, Ralph Grishman, Hoa~Trang Dang, Kira Griffitt, and Joe Ellis.
\newblock 2010.
\newblock Overview of the tac 2010 knowledge base population track.
\newblock In {\em Third Text Analysis Conference (TAC 2010)}.

\bibitem[\protect\citename{Koo \bgroup et al.\egroup }2008]{koo_simple_2008}
Terry Koo, Xavier Carreras, and Michael Collins.
\newblock 2008.
\newblock Simple semi-supervised dependency parsing.
\newblock In {\em Proceedings of {ACL}-08: {HLT}}, pages 595--603, Columbus,
  Ohio, June. Association for Computational Linguistics.

\bibitem[\protect\citename{Li and Ji}2014]{li-ji:2014:P14-1}
Qi~Li and Heng Ji.
\newblock 2014.
\newblock Incremental joint extraction of entity mentions and relations.
\newblock In {\em Proceedings of the 52nd Annual Meeting of the Association for
  Computational Linguistics (Volume 1: Long Papers)}, pages 402--412,
  Baltimore, Maryland, June. Association for Computational Linguistics.

\bibitem[\protect\citename{Liu \bgroup et al.\egroup
  }2015]{liu-EtAl:2015:ACL-IJCNLP}
Yang Liu, Furu Wei, Sujian Li, Heng Ji, Ming Zhou, and Houfeng WANG.
\newblock 2015.
\newblock A dependency-based neural network for relation classification.
\newblock In {\em Proceedings of the 53rd Annual Meeting of the Association for
  Computational Linguistics and the 7th International Joint Conference on
  Natural Language Processing (Volume 2: Short Papers)}, pages 285--290,
  Beijing, China, July. Association for Computational Linguistics.

\bibitem[\protect\citename{Ma \bgroup et al.\egroup
  }2015]{ma-EtAl:2015:ACL-IJCNLP}
Mingbo Ma, Liang Huang, Bowen Zhou, and Bing Xiang.
\newblock 2015.
\newblock Dependency-based convolutional neural networks for sentence
  embedding.
\newblock In {\em Proceedings of the 53rd Annual Meeting of the Association for
  Computational Linguistics and the 7th International Joint Conference on
  Natural Language Processing (Volume 2: Short Papers)}, pages 174--179,
  Beijing, China, July. Association for Computational Linguistics.

\bibitem[\protect\citename{Manning \bgroup et al.\egroup
  }2014]{manning-EtAl:2014:P14-5}
Christopher~D. Manning, Mihai Surdeanu, John Bauer, Jenny Finkel, Steven~J.
  Bethard, and David McClosky.
\newblock 2014.
\newblock The {Stanford} {CoreNLP} natural language processing toolkit.
\newblock In {\em Proceedings of 52nd Annual Meeting of the Association for
  Computational Linguistics: System Demonstrations}, pages 55--60.

\bibitem[\protect\citename{Mikolov \bgroup et al.\egroup
  }2013]{mikolov2013distributed}
Tomas Mikolov, Ilya Sutskever, Kai Chen, Greg Corrado, and Jeffrey Dean.
\newblock 2013.
\newblock Distributed representations of words and phrases and their
  compositionality.
\newblock {\em arXiv preprint arXiv:1310.4546}.

\bibitem[\protect\citename{Miller \bgroup et al.\egroup
  }2004]{miller_name_2004}
Scott Miller, Jethran Guinness, and Alex Zamanian.
\newblock 2004.
\newblock Name tagging with word clusters and discriminative training.
\newblock In Susan Dumais, Daniel Marcu, and Salim Roukos, editors, {\em
  {HLT-NAACL} 2004: Main Proceedings}. Association for Computational
  Linguistics.

\bibitem[\protect\citename{Mitchell \bgroup et al.\egroup
  }2005]{mitchell_ace_2005}
Alexis Mitchell, Stephanie Strassel, Shudong Huang, and Ramez Zakhary.
\newblock 2005.
\newblock Ace 2004 multilingual training corpus.
\newblock {\em Linguistic Data Consortium, Philadelphia}.

\bibitem[\protect\citename{Mnih and Hinton}2007]{mnih_three_2007}
Andriy Mnih and Geoffrey Hinton.
\newblock 2007.
\newblock Three new graphical models for statistical language modelling.
\newblock In {\em Proceedings of the 24th international conference on Machine
  learning}, pages 641--648. {ACM}.

\bibitem[\protect\citename{Nguyen and Grishman}2014]{nguyen_employing_2014}
Thien~Huu Nguyen and Ralph Grishman.
\newblock 2014.
\newblock Employing word representations and regularization for domain
  adaptation of relation extraction.
\newblock In {\em Proceedings of the 52nd Annual Meeting of the Association for
  Computational Linguistics (Volume 2: Short Papers)}, pages 68--74, Baltimore,
  Maryland, June. Association for Computational Linguistics.

\bibitem[\protect\citename{Nguyen and Grishman}2015]{nguyenrelation}
Thien~Huu Nguyen and Ralph Grishman.
\newblock 2015.
\newblock Relation extraction: Perspective from convolutional neural networks.
\newblock In {\em Proceedings of NAACL Workshop on Vector Space Modeling for
  NLP}.

\bibitem[\protect\citename{Nguyen \bgroup et al.\egroup
  }2015]{nguyen-plank-grishman:2015:ACL-IJCNLP}
Thien~Huu Nguyen, Barbara Plank, and Ralph Grishman.
\newblock 2015.
\newblock Semantic representations for domain adaptation: A case study on the
  tree kernel-based method for relation extraction.
\newblock In {\em Proceedings of the 53rd Annual Meeting of the Association for
  Computational Linguistics and the 7th International Joint Conference on
  Natural Language Processing (Volume 1: Long Papers)}, pages 635--644,
  Beijing, China, July. Association for Computational Linguistics.

\bibitem[\protect\citename{Parker \bgroup et al.\egroup
  }2011]{parker2011english}
Robert Parker, David Graff, Junbo Kong, Ke~Chen, and Kazuaki Maeda.
\newblock 2011.
\newblock English gigaword fifth edition, june.
\newblock {\em Linguistic Data Consortium, LDC2011T07}.

\bibitem[\protect\citename{Plank and Moschitti}2013]{plank_embedding_2013}
Barbara Plank and Alessandro Moschitti.
\newblock 2013.
\newblock Embedding semantic similarity in tree kernels for domain adaptation
  of relation extraction.
\newblock In {\em Proceedings of the 51st Annual Meeting of the Association for
  Computational Linguistics (Volume 1: Long Papers)}, pages 1498--1507, Sofia,
  Bulgaria, August. Association for Computational Linguistics.

\bibitem[\protect\citename{Rink and
  Harabagiu}2010]{rink-harabagiu:2010:SemEval}
Bryan Rink and Sanda Harabagiu.
\newblock 2010.
\newblock Utd: Classifying semantic relations by combining lexical and semantic
  resources.
\newblock In {\em Proceedings of the 5th International Workshop on Semantic
  Evaluation}, pages 256--259, Uppsala, Sweden, July. Association for
  Computational Linguistics.

\bibitem[\protect\citename{Roth and Woodsend}2014]{roth_composition_2014}
Michael Roth and Kristian Woodsend.
\newblock 2014.
\newblock Composition of word representations improves semantic role labelling.
\newblock In {\em {EMNLP}}.

\bibitem[\protect\citename{Socher \bgroup et al.\egroup
  }2012]{socher-EtAl:2012:EMNLP-CoNLL}
Richard Socher, Brody Huval, Christopher~D. Manning, and Andrew~Y. Ng.
\newblock 2012.
\newblock Semantic compositionality through recursive matrix-vector spaces.
\newblock In {\em Proceedings of the 2012 Joint Conference on Empirical Methods
  in Natural Language Processing and Computational Natural Language Learning},
  pages 1201--1211, Jeju Island, Korea, July. Association for Computational
  Linguistics.

\bibitem[\protect\citename{Socher \bgroup et al.\egroup
  }2013a]{socher2013parsing}
Richard Socher, John Bauer, Christopher~D Manning, and Andrew~Y Ng.
\newblock 2013a.
\newblock Parsing with compositional vector grammars.
\newblock In {\em In Proceedings of the ACL conference}. Citeseer.

\bibitem[\protect\citename{Socher \bgroup et al.\egroup
  }2013b]{socher-EtAl:2013:EMNLP}
Richard Socher, Alex Perelygin, Jean Wu, Jason Chuang, Christopher~D. Manning,
  Andrew Ng, and Christopher Potts.
\newblock 2013b.
\newblock Recursive deep models for semantic compositionality over a sentiment
  treebank.
\newblock In {\em Empirical Methods in Natural Language Processing}, pages
  1631--1642.

\bibitem[\protect\citename{Sun \bgroup et al.\egroup
  }2011]{sun_semi-supervised_2011}
Ang Sun, Ralph Grishman, and Satoshi Sekine.
\newblock 2011.
\newblock Semi-supervised relation extraction with large-scale word clustering.
\newblock In {\em Proceedings of the 49th Annual Meeting of the Association for
  Computational Linguistics: Human Language Technologies}, pages 521--529,
  Portland, Oregon, {USA}, June. Association for Computational Linguistics.

\bibitem[\protect\citename{Surdeanu \bgroup et al.\egroup
  }2012]{surdeanu2012multi}
Mihai Surdeanu, Julie Tibshirani, Ramesh Nallapati, and Christopher~D Manning.
\newblock 2012.
\newblock Multi-instance multi-label learning for relation extraction.
\newblock In {\em Proceedings of the 2012 Joint Conference on Empirical Methods
  in Natural Language Processing and Computational Natural Language Learning},
  pages 455--465. Association for Computational Linguistics.

\bibitem[\protect\citename{Turian \bgroup et al.\egroup }2010]{turian2010word}
Joseph Turian, Lev Ratinov, and Yoshua Bengio.
\newblock 2010.
\newblock Word representations: a simple and general method for semi-supervised
  learning.
\newblock In {\em Association for Computational Linguistics}, pages 384--394.

\bibitem[\protect\citename{Walker \bgroup et al.\egroup }2006]{walker_ace_2006}
Christopher Walker, Stephanie Strassel, Julie Medero, and Kazuaki Maeda.
\newblock 2006.
\newblock {ACE} 2005 multilingual training corpus.
\newblock {\em Linguistic Data Consortium, Philadelphia}.

\bibitem[\protect\citename{Yu and Dredze}2015]{yu2015learning}
Mo~Yu and Mark Dredze.
\newblock 2015.
\newblock Learning composition models for phrase embeddings.
\newblock {\em Transactions of the Association for Computational Linguistics},
  3:227--242.

\bibitem[\protect\citename{Yu \bgroup et al.\egroup }2015]{yu_combining_2015}
Mo~Yu, Matthew~R. Gormley, and Mark Dredze.
\newblock 2015.
\newblock Combining word embeddings and feature embeddings for fine-grained
  relation extraction.
\newblock In {\em Proceedings of {NAACL}}.

\bibitem[\protect\citename{Zeng \bgroup et al.\egroup
  }2014]{zeng-EtAl:2014:Coling}
Daojian Zeng, Kang Liu, Siwei Lai, Guangyou Zhou, and Jun Zhao.
\newblock 2014.
\newblock Relation classification via convolutional deep neural network.
\newblock In {\em Proceedings of COLING 2014, the 25th International Conference
  on Computational Linguistics: Technical Papers}, pages 2335--2344, Dublin,
  Ireland, August. Dublin City University and Association for Computational
  Linguistics.

\bibitem[\protect\citename{Zhou \bgroup et al.\egroup
  }2005]{zhou_exploring_2005}
GuoDong Zhou, Jian Su, Jie Zhang, and Min Zhang.
\newblock 2005.
\newblock Exploring various knowledge in relation extraction.
\newblock In {\em Association for Computational Linguistics}, pages 427--434.

\end{thebibliography}

\clearpage
\newpage


\section*{Appendix 1: Experiments on ACE 2005 where Gold Entity Types Are Unknown}
\label{sec:appendix}
\paragraph{Experimental Settings:}
For comparison with prior work \cite{plank_embedding_2013}, we (1) generate relation
instances from all pairs of entities within each sentence with three
or fewer intervening entity mentions---labeling those pairs with no
relation as negative instances, (2) use gold entity spans (but not types) at
train and test time, and
(3) evaluate on the 7 coarse relation types, ignoring the subtypes. 
In the training set, 35,990 total relations are annotated of which
only 3,658 are non-nil relations. 
We did not match the number of tokens they reported in the \texttt{cts}
and \texttt{wl} domains. Therefore, in this section we only report the results on the test set
of \texttt{bc} domain. We will leave experiments on additional domains in future work.

We run the same models as in \S\ref{ssec:res_ace} on this task. Here the \fct\ does not use entity type features.
\newcite{plank_embedding_2013} also use Brown clusters and word
vectors learned by latent-semantic analysis (LSA).
In order to make a fair comparison with their method, we also report the \fct\ 
result using Brown clusters (prefixes of length 5) of entity heads as entity types.
Furthermore, we report non-comparable settings using WordNet super-sense tags
of entity heads as types. The WordNet features were also used in their paper but not
as substitution of entity types. 
We use the same toolkit to get the WordNet tags as in \S\ref{sec:exp_setting}.
The Brown clusters are from \cite{koo_simple_2008}\footnote{\url{
http://people.csail.mit.edu/maestro/papers/bllip-clusters.gz}}.

\paragraph{Results:}
Table \ref{tab:pm13_results} shows the results under the low-resource setting.
When no entity types are available, the performance of our \fct\ only model greatly
decreases to 48.15\%, which is consistent with our observation in the ablation tests.
The baseline model also relies heavily on the entity types.
After we remove all the hand-engineering features that contain entity type information,
the performance of our baseline model drop to 40.62\%, even  lower than the 
reduced \fct\ only model.

The combination of baseline model and head embeddings (Baseline + HeadOnly) 
greatly improve the results.
This is consistent with the observation in \newcite{nguyen_employing_2014} that
when the gold entity types are unknown, information of the entity heads
provided by their embeddings will play a more important role.
Combination of the baseline and  \fct\ (Baseline + \fct)
also achieves improvement but not significantly
better than Baseline + HeadOnly.
A possible explanation is that \fct\ becomes less efficient on using
context word embeddings when the entity type information is unavailable.
In this situation the head embeddings provided by \fct\ become the dominating
contribution to the baseline model, making the model have similar behavior
as the Baseline + HeadOnly method.

Finally, we find Brown clusters can help \fct\ when entity types are unknown.
Although the performance is still not significantly better than Baseline + HeadOnly, 
it outperforms all the results in \newcite{plank_embedding_2013} as a \emph{single model},
and with the \emph{same source of features}.
WordNet super-sense tags further improves  \fct{}, and achieves the 
best reported results on this low-resource setting.
These results are encouraging since it shows  \fct\ may be more useful
under the end-to-end setting where predictions of both entity mentions and relation mentions
are required in place of predicting relation based on gold tags \cite{li-ji:2014:P14-1}. 

Recently \newcite{nguyen-plank-grishman:2015:ACL-IJCNLP} proposed a novel way of applying
embeddings to tree-kernels.
From the results, our best single model achieves comparable result with their best single system, while their
combination method is slightly better than ours.
This suggests that we may benefit more from combining the usages of multiple word representations; and we will investigate it in future work.

\begin{table}[htbp]
\centering
\small
\begin{tabular}{|l|c|c|c|c|}
\hline
\multirow{2}{*} & \multicolumn{3}{|c|}{\bf bc} \\
\cline{2-4}
\bf Model & \bf P & \bf R & \bf F1\\
        \hline
       PM'13 (Brown) &54.4&43.4&	48.3\\
       PM'13 (LSA) & 53.9	& 45.2	&49.2\\
       PM'13 (Combination) &55.3&	43.1&	48.5\\
        \hdashline
        (1) \fct\ only& 53.7 & {43.7} & 48.2  \\
        (3) Baseline & 59.4 & 30.9 & 40.6 \\
        (4) + HeadOnly & 64.9 & 41.3 & 50.5 \\
        (5) + \fct & \textbf{65.5} & 41.5 & {50.8}  \\
        \hdashline
        (1) \fct\ only w/ Brown & 64.6 & 40.2 & 49.6\\
        (1) \fct\ only w/WordNet & 64.0 & 43.2 &  51.6\\
       \hline
       Linear+Emb &46.5 & \textbf{49.3} & 47.8\\
       Tree-kernel+Emb (Single) & 57.6 & 46.6 & 51.5\\
       Tree-kernel+Emb (Combination) &58.5 & 47.3 & \textbf{52.3}\\
       \hline
\end{tabular}
\caption{\small{Comparison of models on ACE 2005 out-of-domain test sets
for the low-resource setting, where the
gold entity spans are known but
entity types are unknown.
PM'13 is the results reported in
\newcite{plank_embedding_2013}.
``Linear+Emb" is the implementation of our method (4) in \cite{nguyen-plank-grishman:2015:ACL-IJCNLP}. The ``Tree-kernel+Emb" methods are the enrichments of tree-kernels with embeddings proposed by \newcite{nguyen-plank-grishman:2015:ACL-IJCNLP}.}
}
\label{tab:pm13_results}
\end{table}

\end{document}